\documentclass[runningheads]{llncs}
\usepackage{graphicx}
\usepackage{amsmath,amssymb} 
\usepackage{color}

\usepackage{makecell}
\usepackage{amsmath}
\usepackage{amssymb}
\usepackage{booktabs}
\usepackage{caption}
\usepackage{subcaption}
\usepackage{adjustbox}
\usepackage{xspace}
\usepackage{hyperref}
\usepackage{cleveref}

\begin{document}
\pagestyle{headings}
\mainmatter

\def\ACCV20SubNumber{363}  

\title{Goal-GAN: Multimodal Trajectory Prediction Based on Goal Position Estimation} 
\titlerunning{Goal-GAN}
%
\author{\href{https://orcid.org/0000-0002-4623-8749}{Patrick Dendorfer} \and
\href{https://orcid.org/0000-0001-8105-4737}{Aljo\v{s}a O\v{s}ep} \and
\href{https://orcid.org/0000-0001-8709-1133}{Laura Leal-Taix\'e}}
\authorrunning{P. Dendorfer et al.}
%
\institute{Technical University Munich\\
\email{\{patrick.dendorfer,aljosa.osep,leal.taixe\}@tum.de}}

\maketitle

\newcommand*{\ie}{\emph{i.e.}\@\xspace}
\newcommand*{\etal}{\emph{et al.}\@\xspace}

\newcommand*{\eg}{\emph{e.g.}\@\xspace}

\newcommand*{\cf}{\emph{c.f.}\@\xspace}

\newcommand*{\wrt}{w.r.t. \@\xspace}
\begin{abstract}
In this paper, we present Goal-GAN, an interpretable and end-to-end trainable model for human trajectory prediction. 
Inspired by human navigation, we model the task of trajectory prediction as an intuitive two-stage process: (i) goal estimation, which predicts the most likely target positions of the agent, followed by a (ii) routing module which estimates a set of plausible trajectories that route towards the estimated goal.
We leverage information about the past trajectory and visual context of the scene to estimate a multi-modal probability distribution over the possible goal positions, which is used to sample a potential goal during the inference. 
The routing is governed by a recurrent neural network that reacts to physical constraints in the nearby surroundings and generates feasible paths that route towards the sampled goal. 
Our extensive experimental evaluation shows that our method establishes a new state-of-the-art on several benchmarks while being able to generate a realistic and diverse set of trajectories that conform to physical constraints.
\end{abstract}


\section{Introduction}
\begin{figure}
\begin{center}
\centering
\begin{subfigure}{.32\textwidth}
\includegraphics[width=1\textwidth]{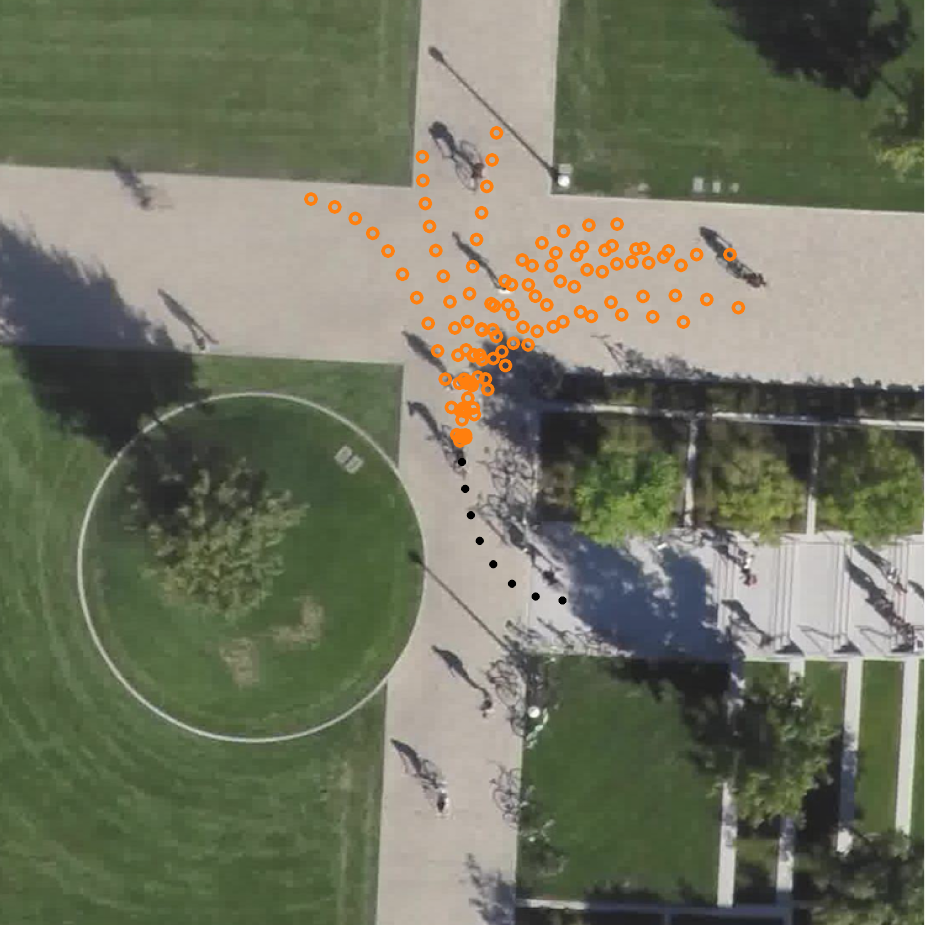}
\caption{Vanilla GAN}\label{fig:VanillaGANIntro}
\end{subfigure}
\begin{subfigure}{.32\textwidth}
\includegraphics[width=1\textwidth]{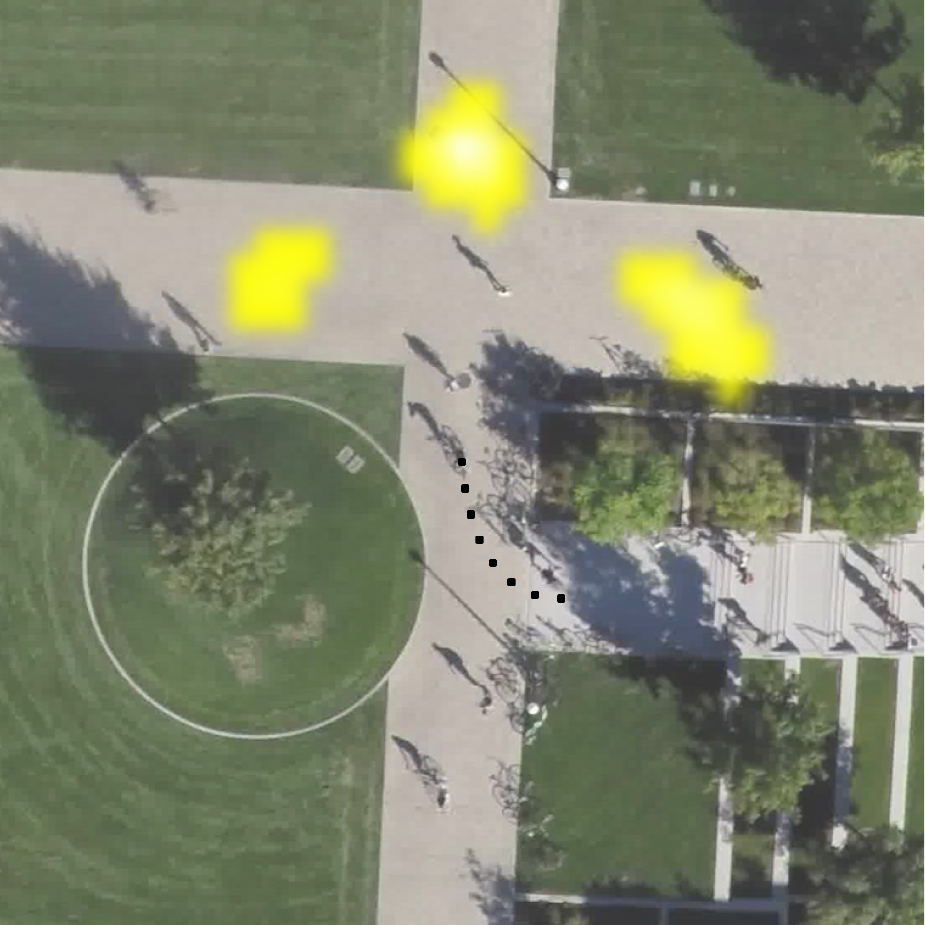}
\caption{Goal Probabilities}\label{fig:GoalProbsIntro}
\end{subfigure}
\begin{subfigure}{.32\textwidth}
\includegraphics[width=1\textwidth]{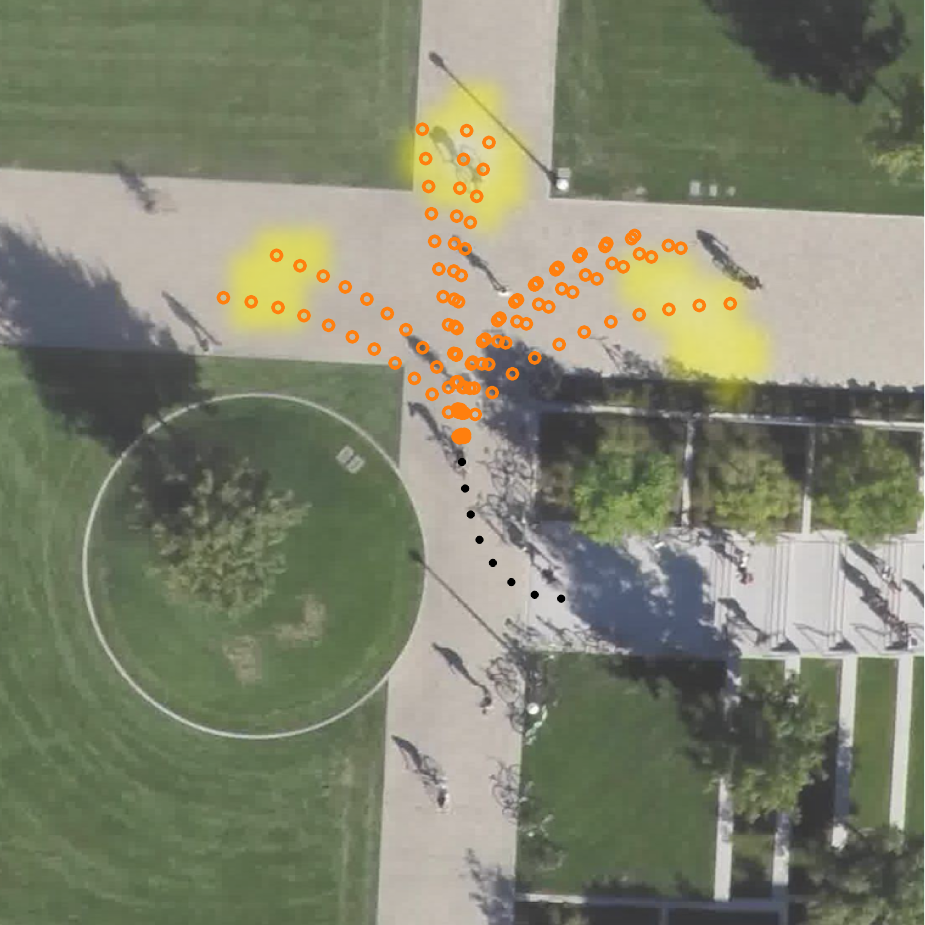}
\caption{Goal-GAN}
\label{fig:GoalGANSamples}
\end{subfigure}
\caption{Visual comparison between predictions of our proposed Goal-GAN and a vanilla GAN. 
In contrast to the baseline, our proposed model covers all three modes and predicts diverse and feasible trajectories by explicitly estimating realistic goals.}\label{fig:TrajectoriesIntro}
\end{center}
\end{figure}

Modeling human motion is indispensable for autonomous systems that operate in public spaces, such as self-driving cars or social robots. 
Safe navigation through crowded scenes and collision prevention requires awareness not only of the present position but also of the future path of all moving objects.
Human trajectory prediction is particularly challenging since pedestrian trajectories depend primarily on their intention -- and the destination of a pedestrian is inherently unknown to the external observer.
Consider the example of a pedestrian reaching a crossroad such as the one depicted in \Cref{fig:TrajectoriesIntro}. Based solely on past observations, we cannot infer the future path of the pedestrian: turning right, left, or going straight, are all equally likely outcomes.

For this reason, a powerful prediction model should be able to capture the {\it multimodality} of this task, \ie, forecast trajectories that cover the distinctive modes present in the scene. 
Furthermore, it should produce a {\it diverse} set of the paths within each mode, reflecting inherent uncertainty in walking style, velocity, and different strategies for obstacle avoidance. 

To capture the stochastic nature of trajectory prediction, state-of-the-art methods leverage generative the power of variational autoencoders (VAEs)~\cite{best-of-many-sampling,desire,Felsen:autoencoder} and/or generative adversarial networks (GANs)~\cite{SGAN,sadeghian2018sophie,kosaraju2019socialbigat} to predict a set of trajectories for every observation. 

While generative methods are widely used to generate diverse outputs, they are unable to explicitly capture the inherent multimodality of pedestrian trajectories. Often, these methods generate highly diverse trajectories but tend to neglect the physical structure of the environment.
The resulting trajectories are not necessarily feasible, and often do not fully cover multiple possible directions that a pedestrian can take (\Cref{fig:VanillaGANIntro}). 
A more natural way of capturing all feasible directions is to first determine an intermediate goal sampled from a distribution of plausible positions, as shown in \Cref{fig:GoalProbsIntro}. In the second step, the model generates the trajectories reaching the sampled positions (\Cref{fig:GoalGANSamples}). While social interactions among agents~\cite{socialLSTM,SGAN,sadeghian2018sophie,kosaraju2019socialbigat}  and local scene interaction have been extensively studied, there are almost no methods tackling the challenge of explicitly learning the inherent multimodal distribution of pedestrian trajectories. 

In this paper, we aim to bridge this gap and explicitly focus on the under-explored problem of generating diverse multimodal trajectories that conform to the physical constraints.
Influenced by recent studies on human navigation~\cite{humanNavigation}
we propose an end-to-end trainable method that separates the task of trajectory prediction into two stages. First, we estimate a posterior over possible goals, taking into account the dynamics of the pedestrian and the visual scene context, followed by the prediction of trajectories that route towards these estimated goals.
Therefore, trajectories generated by our model take both local scene information and past motion of the agent explicitly into account. 
While the estimated distribution of possible goal positions reflects the multimodality in the scene, the routing module reacts to local obstacles and generates diverse and feasible paths. We ensure diversity and realism of the output trajectories by training our network in a generative adversarial setup. 

In summary, our main {\bf contribution} is three-fold: (i) we propose Goal-GAN, a two-stage end-to-end trainable trajectory prediction method inspired by human navigation, which separates the prediction task into goal position estimation and routing. (ii) To this end, we design a novel architecture that explicitly estimates an interpretable probability distribution of future goal positions and allows us to sample from it. Using the Gumbel Softmax trick~\cite{continuousGumbel} enables us to train the network through the stochastic process.
(iii) We establish a new state-of-the-art on several public benchmarks and qualitatively demonstrate that our method predicts realistic end-goal positions together with plausible trajectories that route towards them.
The code for Goal-GAN\footnote{\url{https://github.com/dendorferpatrick/GoalGAN}} is publicly available.

\section{Related Work}

%
Several methods focus on modelling human-human~\cite{SGAN,socialLSTM}, human-space interactions~\cite{ridel2019scene,desire,sadeghian2017carnet}, or both~\cite{sadeghian2018sophie}. 
Recent methods leverage generative models to learn a one-to-many mapping, that is used to sample multimodal future trajectories. 

\noindent\textbf{Trajectory Prediction.} 
Helbing and Molar introduced the Social Force Model (SFM)~\cite{social_force}, a physics-based model, capable of taking agent-agent and agent-space interactions into account. This approach was successfully applied in the domain of multi-object tracking~\cite{scovannericcv2009,pellegriniiccv2009,yamaguchicvpr2011,lealiccv2011}. 
Since then, data-driven models \cite{lealcvpr2014,milan2017online,socialLSTM,stanforddronedataset,SGAN} have vastly outperformed physics-based models. 
Encoder-decoder based methods~\cite{desire,socialLSTM} leverage recurrent neural networks (RNNs)~\cite{rnn-first} to model the temporal evolution of the trajectories with long-short term memory (LSTM) units~\cite{Hochreiter:1997:LSM:1246443.1246450}. 
These deterministic models cannot capture the stochastic nature of the task, as they were trained to minimize the $L_2$ distance between the prediction and the ground truth trajectories. This often results in implausible, average-path trajectories.

Recent methods~\cite{SDDSegmentation,sadeghian2017carnet} focus on human-space interactions using bird-view images~\cite{sadeghian2018sophie} and occupancy grids~\cite{ridel2019scene,hong2019rules} to predict trajectories that respect the structural constraints of the scene. 
Our method similarly leverages bird-eye views. 
However, we use visual information to explicitly estimate feasible and interpretable goal positions, that can, in turn, be used to explicitly sample end-goals that ease the task of future trajectory estimation. 

\noindent\textbf{Generative Models for Trajectory Prediction.}
Recent works~\cite{SGAN,sadeghian2018sophie,kosaraju2019socialbigat} leverage generative models to sample diverse trajectories rather than just predicting a single deterministic output. 
The majority of methods either use variational autoencoders (VAEs)~\cite{VariationalAutoencoder,Felsen:autoencoder,desire,Deo_2018,ivanovic2018trajectron,rhinehart2019precog,sadeghian2017carnet} or generative adversarial networks (GANs)~\cite{GAN,SGAN,sadeghian2018sophie,kosaraju2019socialbigat,amirian2019social}.
Social GAN (S-GAN)~\cite{SGAN} uses a discriminator to learn the distribution of socially plausible paths. Sadeghian~\etal~\cite{sadeghian2018sophie} extend the model to human-environment interactions by introducing a soft-attention~\cite{ShowAttendTell} mechanism. GANs have shown promising results for the task of trajectory prediction, but tend to suffer from mode collapse. 
To encourage the generator to produce more diverse predictions, \cite{best-of-many-sampling} uses a best-of-many sampling approach during training while \cite{kosaraju2019socialbigat} enforces the network to make use of the latent noise vector in combination with BicycleGAN~\cite{toward-image-to-image} based training. 
While producing trajectories with high variance, many trajectories are not realistic, and a clear division between different feasible destinations (reflecting inherent multi-modality of the inherent task) is not clear.
To account for that, we take inspiration from prior work conditioning the trajectory prediction on specific target destinations.

\noindent\textbf{Goal-conditioned forecasting.}
In contrast to the aforementioned generative models that directly learn a one-to-many mapping, 
several methods propose two-stage prediction approaches. 
Similarly to ours, these methods predict first the final (goal) position, followed by trajectory generation that is conditioned on this position.  
Early work of~\cite{Goal-DirectedRehder2015} models the distribution over possible destinations using a particle filter~\cite{Thrun05} while other approaches~\cite{BayesianIntentionGraeme2015} propose a Bayesian framework that estimates both, the destination point together with the trajectory. 
However, these purely probabilistic approaches are highly unstable during training. 
The Conditional Generative Neural System (CGNS)~\cite{li2019conditional} uses variational divergence minimization with soft-attention~\cite{ShowAttendTell} and
\cite{bhattacharyya2019conditional} presents a conditional flow VAE that uses a conditional flow-based prior to effectively structured sequence prediction. 
These models condition their trajectory generator on initially estimated latent codes but do not explicitly predict a goal distribution nor sample an explicit goal position. Most recently, \cite{deo2020trajectory} proposes P2TIRL that uses a maximum entropy inverse reinforcement learning policy to infer goal and trajectory plan over a discrete grid. P2TRL assigns rewards to future goals that are learned by the training policy which is slow and computationally expensive. In contrast, we directly learn the multimodal distribution over possible goals using a binary cross-entropy loss between the (discrete) probability distribution estimate and the ground truth goal position. This makes our work the first method (to the best of our knowledge) that directly predicts an explicit (and discrete) probability distribution for multimodal goals and is efficiently end-to-end trainable. 

\section{Problem Definition}

We tackle the task of predicting the future positions of pedestrians, parametrized via $x$ and $y$ coordinates in the 2D ground plane. As input, we are given their past trajectory and visual information of the scene, captured from a bird-view.

We observe the trajectories $X_i = \{ \left(x^t_i, y^t_i \right)\in \mathbb{R}^2 \vert t\,=\,1, \dots, t_{obs}$ \} of $N$ currently visible pedestrians and a top-down image $I$ of the scene, observed at the timestep $t_{obs}$.
Our goal is to predict the future positions $Y_i = \{ \left(x^t_i, y^t_i \right) \in \mathbb{R}^2 \vert t\,=\,t_{obs}+ 1, \dots, t_{pred}\}$. 

In the dataset, we are only given one future path for $t_{obs}$ -- in particular, the one that was observed in practice. We note that multiple distinctive trajectories could be realist for this observed input trajectory.
Our goal is, given the input past trajectory $X_i$, to generate $k \in \{1, \dots, K \} $  multiple future samples $\hat{Y}^k_i$ for all pedestrians $i \in \{1, \dots, N\}$. These should cover all feasible modes and be compliant with the physical constraints of the scene.


\begin{figure*}[ht]
\begin{center}
   \includegraphics[width=1\linewidth]{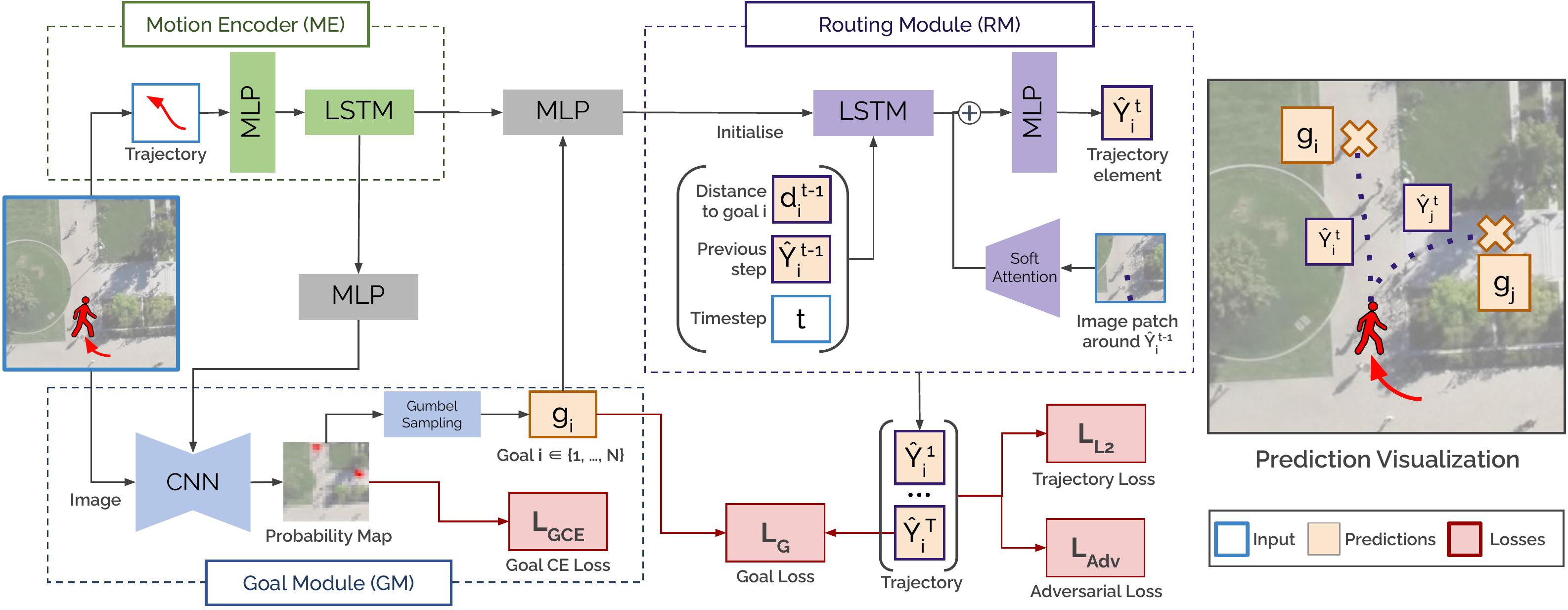}
\end{center}
   \caption{\textbf{Overview of model architecture: } Our model consists of three components: 1) Motion Encoder, 2) Goal Module, and the 3) Routing Module. The Goal Module combines the dynamic features of the  Motion Encoder and the scene information to predict a final goal $g$. The Routing Module generates the future trajectory while taking into account the dynamic features and the estimated goal. During inference, we generate multiple trajectories $i$ by sampling goals from the estimated goal probability map.}
\label{fig1:model_overview}
\end{figure*}

\section{Goal-GAN}
When pedestrians walk through public spaces, they aim to reach a predetermined goal~\cite{humanNavigation}, which depends on their intentions and the scene context. Once the goal is set, humans route to their final destination while reacting to obstacles or other pedestrians along their way. 
This observation motivates us to propose a novel two-stage architecture for trajectory prediction that first estimates the end-goal position and then generates a trajectory towards the estimated goal. 
Our proposed Goal-GAN consists of three key components, as shown in \Cref{fig1:model_overview}. 
\begin{itemize}
\item {\it Motion Encoder (ME):} extracts the pedestrians' dynamic features recursively with a long short-term memory (LSTM) unit capturing the speed and direction of motion of the past trajectory. 
\item {\it Goal Module (GM):} combines visual scene information and dynamic pedestrian features to predict the goal position for a given pedestrian. This module estimates the probability distribution over possible goal (target) positions, which is in turn used to sample goal positions. 
\item {\it Routing Module (RM):} generates the trajectory to the goal position sampled from the GM. While the goal position of the prediction is determined by the GM, the RM generates feasible paths to the predetermined goal and reacts to obstacles along the way by using visual attention.
\end{itemize}

\Cref{fig1:model_overview} shows an overview of our model. In the following sections, we motivate and describe the different components in detail.

\subsection{Motion Encoder (ME)}

The past trajectory of a pedestrian is encoded into the Motion Encoder (ME), which serves as a dynamic feature extractor to capture the speed and direction of the pedestrian, similarly to~\cite{socialLSTM,SGAN}. 
Each trajectory's relative displacement vectors  $\left( \Delta x_i^t, \Delta y_i^t \right)$ are embedded into a higher dimensional vector $e^t$ with a multi-layer perceptron (MLP). The output is then fed into an LSTM, which is used to encode the trajectories.
The hidden state of the LSTM, $h_{M\!E}$, is used by the other modules to predict the goal and to decode the trajectory for each pedestrian. 

\subsection{Goal Module (GM)}
\begin{figure}
 
\begin{center}
   \includegraphics[width=1\linewidth]{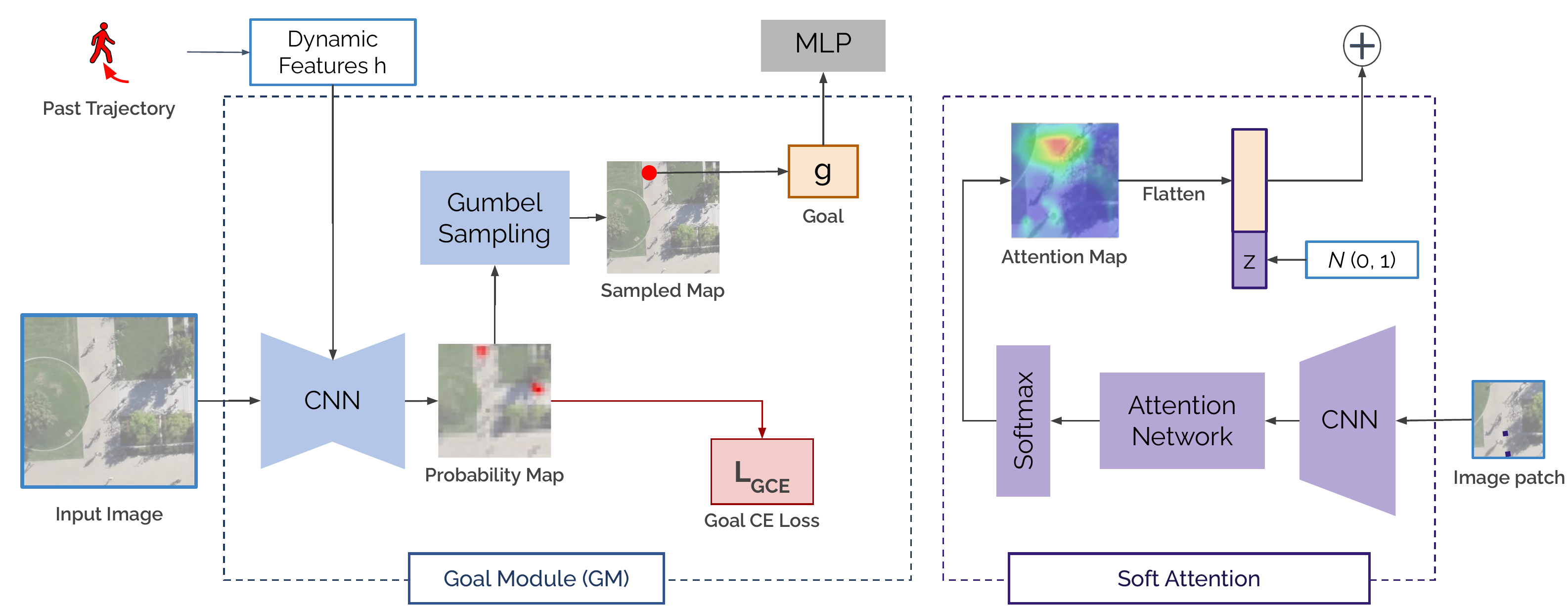}
\end{center}
   \caption{\textbf{Goal Module (GM) and Soft Attention (SA).} The Goal Module samples a goal coordinate $g$ while the soft attention assigns attention scores to spatial positions.}
\label{fig:goalsamplingvsattention}
\end{figure}
In our work, we propose a novel Goal Module (GM).  The Goal Module combines visual and dynamic features of the pedestrian to estimate a distribution of possible end goals. As can be seen in \Cref{fig:TrajectoriesIntro}, the scene dictates the distinctive modes for possible trajectories. Here, the pedestrian can go left, right, or straight. 
The Goal Module is responsible for capturing all the possible modes and predicting a final goal position, \ie, choosing one of the three options.

\noindent{\bf Architecture.}
In order to estimate the goal distribution, the network assesses the visual scene and the dynamics of the pedestrian. 
The visual scene is represented as an RGB image (or a semantic map) of size $H\times W$, captured from a bird-eye view. This image is input to the goal module.

The scene image is passed through an encoder-decoder CNN network with skip connections similar to~\cite{Ronneberger2015UNetCN}. Before the decoder, the scene image features in the bottleneck layer are concatenated with the motion features $h_{M\!E}$ from the Motion Encoder.
Intuitively, the CNN decoder should analyze both the past trajectory and the scene to estimate the future target positions -- goals. The module outputs a probability distribution that reflects the multimodal directions for a given input trajectory and scene.

\noindent{\bf Training through sampling.} \label{sec:gumbelsoftmax}
The CNN decoder outputs a score map $\mathbf{\alpha} = \left( \alpha_1, \alpha_2, \dots , \alpha_n \right)$ for which each  $\alpha_i$ reflects the probability of a particular cell being the end-goal location of the agent. 

The discrete probability distribution $\alpha$ is used to sample an end-goal by using the Gumbel-Softmax-Trick~\cite{continuousGumbel}. This allows us to sample a discrete distribution over possible goal locations while being able to backpropagate the loss through the stochastic process.
The resulting two-dimensional goal position $g$ is sampled randomly from the 2D grid representing the scene, 

\noindent{\bf Goal Sampling vs. Soft Attention.}
A major novelty of our work is the Goal Module that replaces soft attention~\cite{ShowAttendTell} to process the scene contextual information~\cite{sadeghian2018sophie,sadeghian2017carnet}. 
Both approaches are illustrated in \Cref{fig:goalsamplingvsattention}.
A soft attention module assigns attention scores to spatially relevant positions based on the visual CNN features. In~\cite{sadeghian2018sophie}, the attention values are combined with random noise and fed to the trajectory decoder to generate scene-aware multimodal trajectories. However, this often leads to unsatisfying results when the network simply ignores the spatial attention scores or has difficulties combining the attention values with the noise to capture all modes in the scene. 

We argue that the attention module is useful when predicting the route towards a goal (as we show in \Cref{sec:RM}), as it encourages the feasibility of the predicted trajectories. However, the model that solely relies on soft visual attention mechanism tends to generate trajectories that do not capture the multimodal nature of the task, as illustrated in \Cref{fig:TrajectoriesIntro}.
Furthermore, in \Cref{sec:experiments}, we experimentally confirm that stochasticity of the task is reflected better when sampling from the learned probability distribution, produced by our Goal Module, compared to merely relying on noise injection.

We can directly train the module for the goal position estimation using the Gumbel Softmax trick~\cite{continuousGumbel}, in combination with the standard cross-entropy loss, which is directly applied to the estimated goal distribution based on the observed (final) position of the ground truth trajectories. 
We emphasize that we do not use nor need any other data than what is provided in the standard training set. 

During the inference, we simply sample the goal from the learned probability distribution and pass it to the decoder. This significantly eases the task for the decoder, as the Goal Module already assesses the visual surroundings and only passes a low dimensional input into the routing module. 

\subsection{Routing Module (RM)}
\label{sec:RM}
The Routing Module (see \Cref{fig1:model_overview}) is the third component of our method. It combines the dynamic features and the global goal estimate to generate the final trajectory prediction. 
The RM consists of an LSTM network, a visual soft attention network (ATT), and an additional MLP layer that combines the attention map with the output of the LSTM iteratively at each timestep. 

First, we forward the goal estimate embedding $e_g$ and the object dynamics embedding $h_{M\!E}$ (given by the motion encoder, ME) to an MLP to initialise the hidden state $h^0_{RM}$ of the RM. 

Then, we recursively estimate predictions for the future time steps. To this end, the LSTM in the RM obtains three inputs: the previous step prediction $\hat{Y}^{t-1}$, the remaining distance to the estimated goal $d_{t-1} = g - \hat{Y}^{t-1}$ and the current scalar timestep value $t$. 

To assess the traversability of the local surroundings, we apply soft attention~\cite{ShowAttendTell} on the image patch centered around the current position of the pedestrian. 
As shown in the \Cref{fig:goalsamplingvsattention}, we combine the output of the LSTM with the attention map $F^t$ to predict the next step $\hat{Y}^t$. 
The visual attention mechanism allows the RM to react to obstacles or other nearby structures. Finally, we use both the dynamic and visual features to predict the final prediction $\hat{Y}^t$. 

\subsection{Generative Adversarial Training}
In our work, we use a Generative Adversarial Network (GAN) to train our trajectory generator to output realistic and physically feasible trajectories. The GAN consists of a Generator and Discriminator network competing in a two-player min-max game. While the generator aims at producing feasible trajectories, the discriminator learns to differentiate between real and fake samples, i.e., feasible and unfeasible trajectories. 
Adversarial training is necessary because, in contrast to prediction accuracy, it is not possible to formulate a differential loss in a closed mathematical form that captures the concept of feasibility and realism of the generated trajectories. 

The discriminator network consists of an LSTM network that encodes the observed trajectory $X$. This encoding is used to initialize the second LSTM that processes the predicted trajectory $Y$ together with visual features (obtained from the CNN network, that encodes the image patch centered around the current position) at each time step. 
Finally, the last hidden state of the $\text{LSTM}_{pred}$ is used for the final output of the discriminator.

\subsection{Losses}
For training our Goal-GAN we use multiple losses addressing the different modules of our model. 
To encourage the generator to predict trajectories, that are closely resembling the ground truth trajectories, we use a best-of-many~\cite{best-of-many-sampling} distance loss $\mathcal{L}_{L2} = \min_k \lVert Y - \hat{Y}^{(k)} \rVert_2$  between our predictions $\hat{Y}$ and the ground truth $Y$. 
As an adversarial loss, we employ the \textit{lsgan}~\cite{mao2016squares} loss: 
\begin{equation}
\mathcal{L}_{Adv} = \frac{1}{2} \mathbb{E} \, [\left(D\left(X, Y\right) - 1\right)^2] + \frac{1}{2} \mathbb{E}\, [ D (X, \hat{Y} )^2], 
\end{equation}
due to the fact, the original formulation~\cite{GAN} using a classifier with sigmoid cross-entropy function potentially leads to the vanishing gradient problem. 

To encourage the network to take into account the estimated goal positions for the prediction, we propose a goal achievement losses $\mathcal{L}_{G}$ that measures the $L_2$ distance between the goal prediction $g$  and the actual output $\hat{Y}^{t_{pred}}$, 
\begin{equation}
\mathcal{L}_{G} = \lVert g - \hat{Y}^{t_{pred}} \rVert_2.
\end{equation}
In addition, we use a cross-entropy loss
\begin{equation} 
\mathcal{L}_{GCE} = - \log \left( p_i \right),
\end{equation}
where $p_i$ is the probability that is predicted from the Goal Module for the grid cell $i$ corresponding to the final ground-truth position.
The overall loss is the combination of the partial losses weighted by $\lambda$: 
\begin{equation}
\begin{aligned}
& \mathcal{L}  =\lambda_{Adv} \, \mathcal{L}_{Adv} +  \mathcal{L}_{L2} + \lambda_{G} \,\mathcal{L}_{G} +  \lambda_{GCE } \, \mathcal{L}_{GCE} .
\end{aligned}
\end{equation}

\section{Experimental Evaluation}\label{sec:experiments}

In this section, we evaluate our proposed Goal-GAN on three standard datasets used to assess the performance of pedestrian trajectory prediction models: \newline 
ETH~\cite{ETH-data}, UCY~\cite{UCY-data} and Stanford Drone Dataset (SDD)~\cite{stanforddronedataset}. 
To assess how well our prediction model can cover different possible modes (splitting future paths), we introduce a new, synthetically generated scene. 

We compare our method with several state-of-the-art methods for pedestrian trajectory prediction and we qualitatively demonstrate that our method produces multi-modal, diverse, feasible, and interpretable results.

\noindent\textbf{Evaluation measures.}
We follow the standard evaluation protocol and report the prediction accuracy using Average Displacement Error (ADE) and Final Displacement Error (FDE). 
Both measures are computed using the $L_2$ distance between the prediction and ground truth trajectories.
The generative models are tested on these metrics with a $N-K$ variety loss~\cite{best-of-many-sampling,SGAN,sadeghian2018sophie}. As in the previous work~\cite{stanforddronedataset,socialLSTM}, we observe 8 time steps (3.2 seconds) and predict the future 12 time steps (4.8 seconds) simultaneously for all pedestrians in the scene. 

\noindent{\bf Visual input and coordinates.}
As in~\cite{sadeghian2018sophie}, we use a single static image to predict trajectories in a given scene. 
We transform all images into a top-down view using the homography transformation provided by the respective datasets. This allows us to perform all predictions in real-world coordinates. 


\subsection{Benchmark Results}
\begin{table}
\centering
\caption{Quantitative results for ETH~\cite{ETH-data} and UCY~\cite{UCY-data} of Goal-GAN and baseline models predicting 12 future timesteps. We report ADE and FDE in meters.}\label{biwi}
\begin{adjustbox}{width=\columnwidth,center}
\begin{tabular}{l*{7}{c}}\toprule
\multicolumn{1}{c}{}&\multicolumn{6}{c}{Baseline}&\multicolumn{1}{c}{ Ours}\\
 \cmidrule(lr){2-7} \cmidrule(l){8-8} 
 Dataset& \makecell{ S-LSTM\\\cite{socialLSTM}}&\makecell{ S-GAN\\\cite{SGAN}}&\makecell{ S-GAN-P\\
\cite{SGAN}}&\makecell{ SoPhie\\\cite{sadeghian2018sophie}}&\makecell{ S-BiGAT\\\cite{kosaraju2019socialbigat}}& \makecell{ CGNS\\\cite{li2019conditional}}&\makecell{ Goal\\ GAN}\\
\midrule
K&  1 & 20 & 20 & 20 & 20 & 20 & 20 \\ \midrule
\textbf{ETH}&1.09/2.35&0.81/1.52&0.87/1.62&0.70/1.43&0.69/1.29& 0.62/1.40 & \textbf{0.59}/\textbf{1.18}\\
\textbf{HOTEL}&0.79/1.76&0.72/1.61&0.67/1.37&0.76/1.67&0.49/1.01 & 0.70/0.93 & \textbf{0.19}/\textbf{0.35}\\
\textbf{UNIV}&0.67/1.40&0.60/1.26&0.76/1.52&0.54/1.24&0.55/1.32 &\textbf{0.48}/1.22&0.60/\textbf{1.19}\\
\textbf{ZARA1}&0.47/1.00&0.34/0.69&0.35/0.68&\textbf{0.30}/0.63&\textbf{0.30}/0.62 & 0.32/\textbf{0.59}&0.43/0.87\\
\textbf{ZARA2}&0.56/1.17&0.42/0.84&0.42/0.84&0.38/0.78&0.36/0.75 & 0.35/0.71& \textbf{0.32}/\textbf{0.65}\\
\hline\hline\textbf{AVG}&0.72/1.54&0.58/1.18&0.61/1.21&0.54/1.15& 0.48/1.00 & 0.49/0.97 & \textbf{0.43}/\textbf{0.85}\\
\bottomrule\end{tabular}


\end{adjustbox}
\end{table}
In this section, we compare and discuss our method’s performance against state-of-the-art on ETH~\cite{ETH-data}, UCY~\cite{UCY-data} and SDD~\cite{stanforddronedataset} datasets.\newline
\noindent\textbf{Datasets. } ETH~\cite{ETH-data} and UCY datasets~\cite{UCY-data} contain 5 sequences (ETH:2, UCY: 3), recorded in 4 different scenarios. All pedestrian trajectories are converted into real-world coordinates and interpolated to obtain positions every 0.4 seconds. For training and testing, we follow the standard leave-one-out approach, where we train on 4 datasets and test on the remaining one. 
The Stanford Drone Dataset (SDD)~\cite{stanforddronedataset} consists of $20$ unique video sequences captured at the Stanford University campus. The scenes have various landmarks such as roundabouts, crossroads, streets, and sidewalks, which influence the paths of pedestrians. 
In our experiments, we follow the train-test-split of~\cite{trajnet} and focus on pedestrians.\newline
\noindent{\bf Baselines.} We compare our model to several published methods. 
S-LSTM~\cite{socialLSTM} uses a LSTM encoder-decoder network with social pooling. S-GAN~\cite{SGAN} leverages a GAN framework and S-GAN-P~\cite{SGAN} uses max-pooling to model social interactions. 
SoPhie~\cite{sadeghian2018sophie} extends the S-GAN model with a visual and social attention module, and {Social-BiGAT}~\cite{kosaraju2019socialbigat} uses a BicycleGAN~\cite{image-to-image} based training. %
DESIRE~\cite{desire} is an inverse optimal control based model, that utilizes generative modeling. CARNet~\cite{sadeghian2017carnet} is a physically attentive model.
The Conditional Generative Neural System (CGNS)~\cite{li2019conditional} uses conditional latent space learning with variational divergence minimization to learn feasible regions to produce trajectories.  CF-VAE~\cite{bhattacharyya2019conditional} leverages a conditional normalizing flow-based VAE and P2TIRL~\cite{deo2020trajectory} uses a grid-based policy learned with
maximum entropy inverse reinforcement learning policy. 
As none of the aforementioned provide publicly available implementation, we outline the results reported in the respective publications.

\noindent\textbf{ETH and UCY. }
We observe a clear trend -- the generative models improve the performance of the deterministic approaches, as they are capable of sampling a diverse set of trajectories. Compared to other generative models, Goal-GAN achieves state-of-the-art performance with an overall decrease of the error of nearly $15\%$ compared to \textit{S-BiGAT} and \textit{CGNS}. 
While \textit{SoPhie} and \textit{S-BiGAT} also use visual input, these models are unable to effectively leverage this information to discover the dominant modes for the trajectory prediction task, thus yielding a higher prediction error. It has to be pointed out that Goal-GAN decreases the average FDE by $0.12m$ compared to the current state-of-the-art method. We explain the drastic increase in performance with our new Goal Module as we can cover the distribution of all plausible modes and are therefore able to generate trajectories lying close to the ground truth. \newline
\noindent\textbf{Stanford Drone Dataset. }
We compare our model against other baseline methods on the SDD and report ADE and FDE in pixel space. 
As it can be seen in \Cref{Table:SDD}, Goal-GAN achieves state-of-the-art results on both metrics, ADE and FDE. Comparing Goal-GAN against the best non-goal-conditioned method, \textit{SoPhie}, Goal-GAN decreases the error by $25\%$. This result shows clearly the merit of having a two-stage process of predicting a goal estimate over standard generator methods using only soft attention modules but does not explicitly condition their model on a future goal. Further, it can be understood that multimodal trajectory predictions play a major role in the scenes of the SDD. Also, Goal-GAN exceeds all other goal-conditioned methods and is on par with P2TIRL  (which was not yet published during the preparation of this work).
\begin{table}
\centering
\caption{Quantitative results for the Stanford Drone Dataset (SDD)~\cite{stanforddronedataset} of Goal-GAN and baseline models predicting 12 future timesteps. We report ADE and FDE in pixels. }\label{Table:SDD}
\begin{adjustbox}{width=\columnwidth,center}
\begin{tabular}{lccccccccc} \toprule
\multicolumn{1}{c}{}&\multicolumn{8}{c}{Baseline}&\multicolumn{1}{c}{ Ours}\\
\cmidrule(lr){2-9} \cmidrule(l){10-10} 
& \makecell{S-LSTM\\\cite{socialLSTM}}&\makecell{S-GAN\\\cite{SGAN}}&\makecell{CAR-NET\\\cite{sadeghian2017carnet}}&\makecell{DESIRE\\\cite{desire}}&\makecell{SoPhie\\\cite{sadeghian2018sophie}}&\makecell{CGNS\\\cite{li2019conditional}}&\makecell{CF-VAE\\\cite{bhattacharyya2019conditional}}&\makecell{P2TIRL\\\cite{deo2020trajectory}}&\makecell{Goal\\GAN}\\ \midrule
K & 1 & 20 & 20 & 20 & 20 & 20 & 20 & 20 & 20 \\ \midrule
\textbf{ADE} &57.0 &27.3&25.7 &19.3&16.3 & 15.6 & 12.6 & 12.6 &  \textbf{12.2}  \\ 
\textbf{FDE} &31.2 &41.4 &51.8 &34.1 &29.4 & 28.2 & 22.3 & \textbf{22.1} & \textbf{ 22.1 } \\ 
\bottomrule \end{tabular}
\end{adjustbox}
\end{table}
%

 
\subsection{Assessing Multimodality of Predictions on Synthetic Dataset}
In this section, we conduct an additional experiment using synthetically generated scenarios to study the multimodality of the predictions. 
We compare the performance of Goal-GAN against two vanilla GAN baselines, with and without visual soft attention~\cite{ShowAttendTell}. 
The synthetic dataset allows us to explicitly control multimodality and feasibility of the (generated) ground truth trajectories, as the other datasets do not provide that information. 
\begin{table}
\centering
\caption{Quantitative results on our synthetic dataset. We show results obtained by a GAN baseline~\cite{SGAN} and different versions of our Goal-GAN model, predicting 12 future time steps. We report ADE, FDE, F (feasibility) and MC (mode coverage) for $k=10$ sampled trajectories for each scene. We also report the negative log-likelihood (NLL) of the ground truth trajectory computed with the KDE (Kernel Density Estimate), following \cite{ivanovic2018trajectron}.
}\label{Table:SyntheticDataset}
\begin{adjustbox}{width=\columnwidth,center}

\begin{tabular}{llccccc}\toprule
Model &	Loss &	ADE $\downarrow$	& FDE $\downarrow$ &	F $\uparrow$ &	MC $\uparrow$ &	NLL $\downarrow$ \\ \midrule
GAN w/o visual	& $\mathcal{L}_{L2} + \mathcal{L}_{Adv}$	&0.70&	1.49&	59.94&	78.51	&4.54\\
GAN w visual&	$\mathcal{L}_{L2} + \mathcal{L}_{Adv}$	&0.68&	1.27&	66.51&	85.12&	4.47 \\ \midrule
Goal-GAN & $\phantom{i \mathcal{L}_{L2} + \mathcal{L}_{Adv} +  } \mathcal{L}_{GCE} + \mathcal{L}_{G}$ & 2.09 & 1.27 & 	76.78	& 88.22 & 	\textbf{3.76} \\
Goal-GAN & $\mathcal{L}_{L2} + \phantom{ i\mathcal{L}_{Adv} +  } \mathcal{L}_{GCE} + \mathcal{L}_{G}$ & 0.62 & 1.20 & 85.05 &89.27 &	3.90 \\
Goal-GAN w/o GST	& $\mathcal{L}_{L2} + \mathcal{L}_{Adv} +  \mathcal{L}_{GCE} + \mathcal{L}_{G}$	& 0.84 & 1.45 & 76.84 & 86.27 & 4.18 \\ \midrule
Goal-GAN	(full model) &$\mathcal{L}_{L2} + \mathcal{L}_{Adv} +  \mathcal{L}_{GCE} + \mathcal{L}_{G}$ &\textbf{ 0.55} & \textbf{1.01} & \textbf{89.47} & \textbf{92.48 }& 3.88 \\
\bottomrule
\end{tabular}

\end{adjustbox}
\end{table}

\noindent\textbf{Dataset.} We generate trajectories using the Social Force Model~\cite{social_force} in the \textit{hyang 4} scene of the SDD dataset~\cite{stanforddronedataset}. 
To ensure the feasibility of the generated trajectories, we use a two-class (manually labeled) semantic map, that distinguishes between feasible (walking paths) from unfeasible (grass) areas. 
We simulate $250$ trajectories approaching and passing the two crossroads in the scene.

\noindent\textbf{Additional Evaluation Measures.}
In addition to ADE and FDE, we follow~\cite{ivanovic2018trajectron,thiede2019analyzing} to measure the multimodality of the distribution of generated trajectories. Here we evaluate the negative log-likelihood (NLL) of the ground truth trajectories using a Kernel Density Estimate (KDE) from the sampled trajectories at each prediction timestep.
In addition, we define a new mode coverage ({MC}) metric. For each scene, MC assesses if at least one of the $k$ generated trajectories $\hat{y}$ reaches the final position of the ground truth final up to a distance of $2m$: 
\begin{equation}
\text{MC} = \frac{1}{n} \sum_i^n S\left(\mathbf{\hat{y}}_i \right) \text{ with }
S\left(\mathbf{\hat{y}}  \right) = \begin{cases} 
    1 \quad \text{if} \, \,  \exists k, \, \, \lVert \hat{y}^k - y \rVert_2 < 2m \\
      0 \quad \text{else}.
   \end{cases}
\end{equation} 
To evaluate the feasibility of the trajectories, we report the ratio of trajectories lying inside the feasible area $\mathcal{F}$, \ie, predictions staying on the path:
\begin{equation}
\text{F} = \frac{1}{n} \sum_{i, k}^n f\left(\hat{y}_i^k\right) \text{ with }
f\left(\hat{y}  \right) = \begin{cases} 
1 \quad \text{if} \, \,  \hat{y} \in \mathcal{F} \\
  0 \quad \text{else}.
\end{cases}
\end{equation}
\noindent\textbf{Results.} 
As can be seen in \Cref{Table:SyntheticDataset}, the vanilla GAN baseline~\cite{SGAN} that is not given access to the visual information, yields ADE/FDE of $0.70/1.49$, respectively. 
Adding visual information yields a performance boost ($0.68/1.27$), however, it is still not able to generate multimodal and feasible paths.
When we add our proposed goal module (Goal-GAN) and train it using our full loss, we observe a large boost of performance \wrt multimodality ($7.36$ increase in terms of MC) and feasibility ($10.26$ increase in terms of F). 
To ablate our model, we train our network using different loss components, incentivizing the network to train different modules of the network. 
A variant of our model, trained only with goal achievement loss $\mathcal{L}_{G}$ and adversarial loss $\mathcal{L}_{Adv}$ can already learn to produce multimodal trajectories (MC of $88.22$), however, yields a high ADE error of $2.09$. 
The addition of $L_2$ loss $\mathcal{L}_{L2}$ significantly increases the accuracy of the predictions ($1.47$ reduction in ADE), and at the same time, increases the quality and feasibility ($8.26$ increase in F), of the predictions. 
This confirms that our proposed goal module, which explicitly models the distribution over the future goals, is vital for accurate and realistic predictions. 
Furthermore, we note that the performance drastically drops if we train the full model without the Gumbel-Softmax Trick (GST) (see \Cref{sec:gumbelsoftmax}) which seems to be crucial for stable training, enabling the loss back-propagation through the stochastic sampling process in the Goal Module. 

\begin{figure}[ht]
\centering
\begin{subfigure}{1\textwidth}
\includegraphics[width=0.23\textwidth]{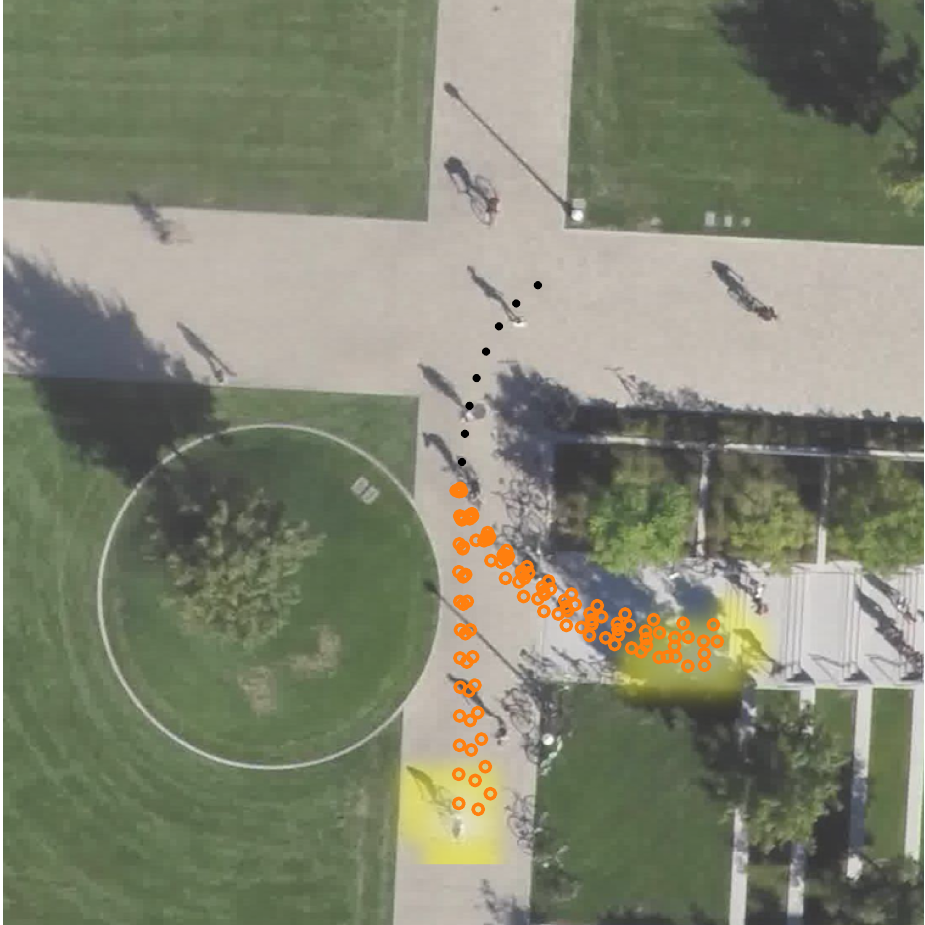} \hfill
\includegraphics[width=0.23\textwidth]{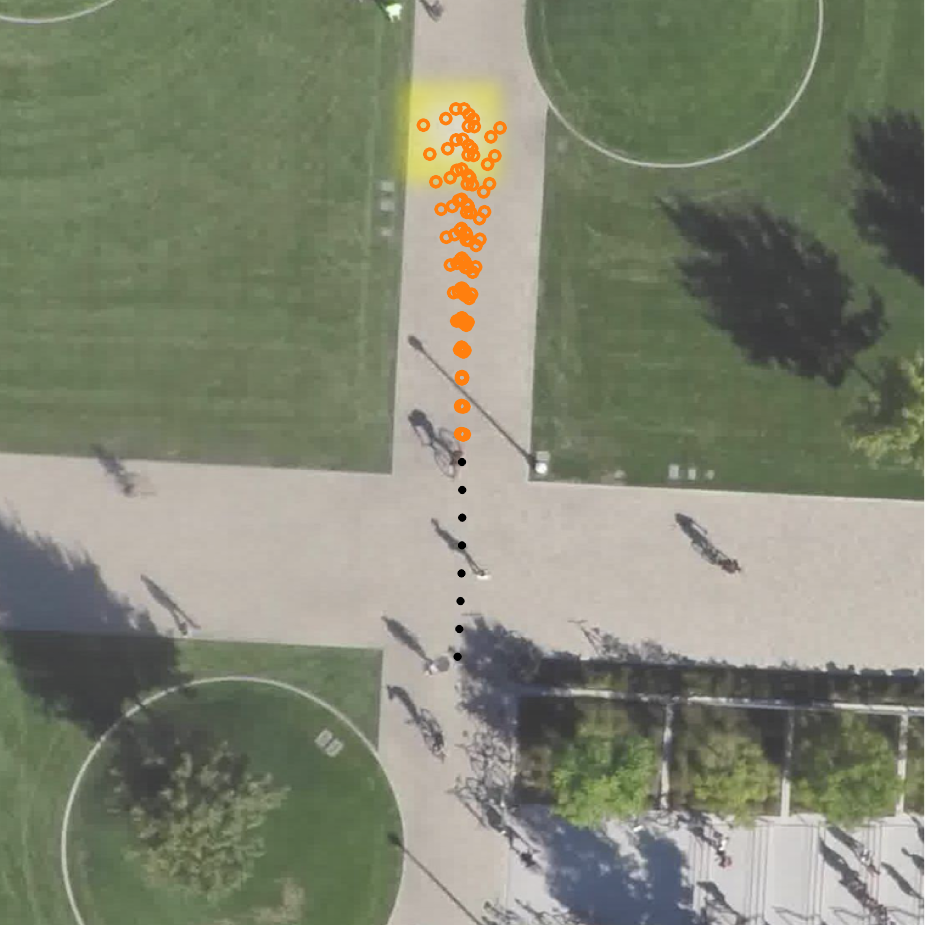}
\hfill
\includegraphics[width=0.23\textwidth]{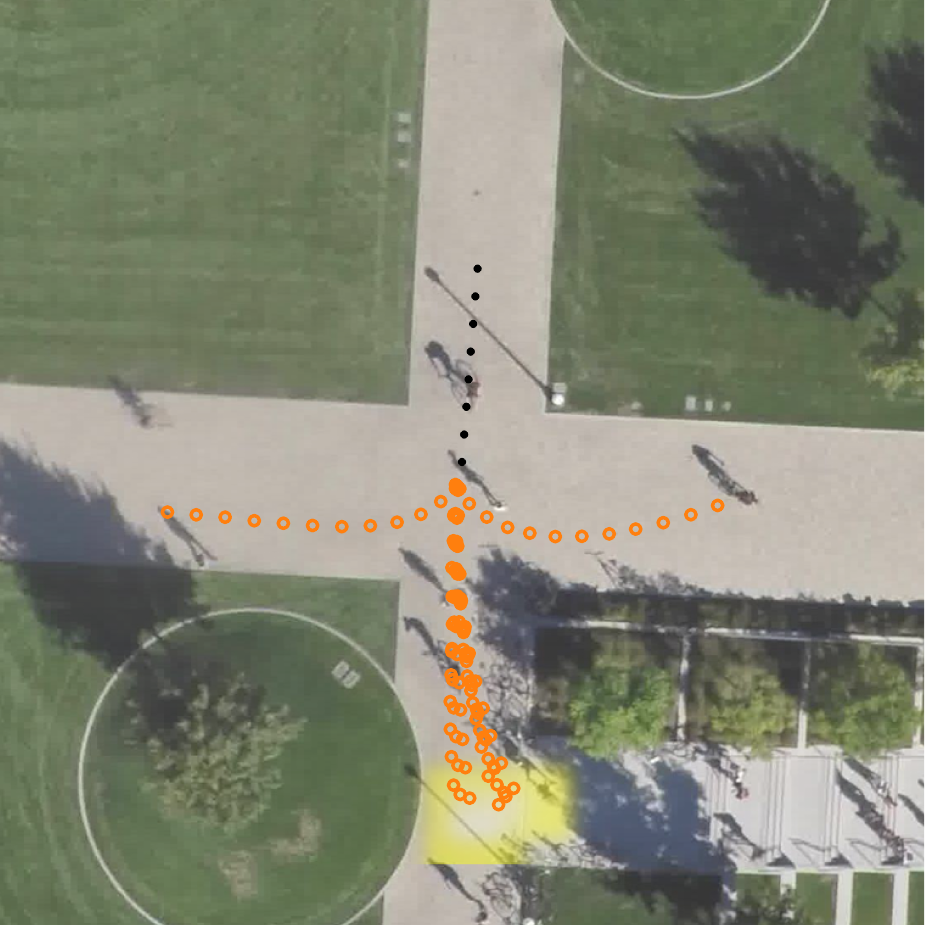}
\hfill
\includegraphics[width=0.23\textwidth]{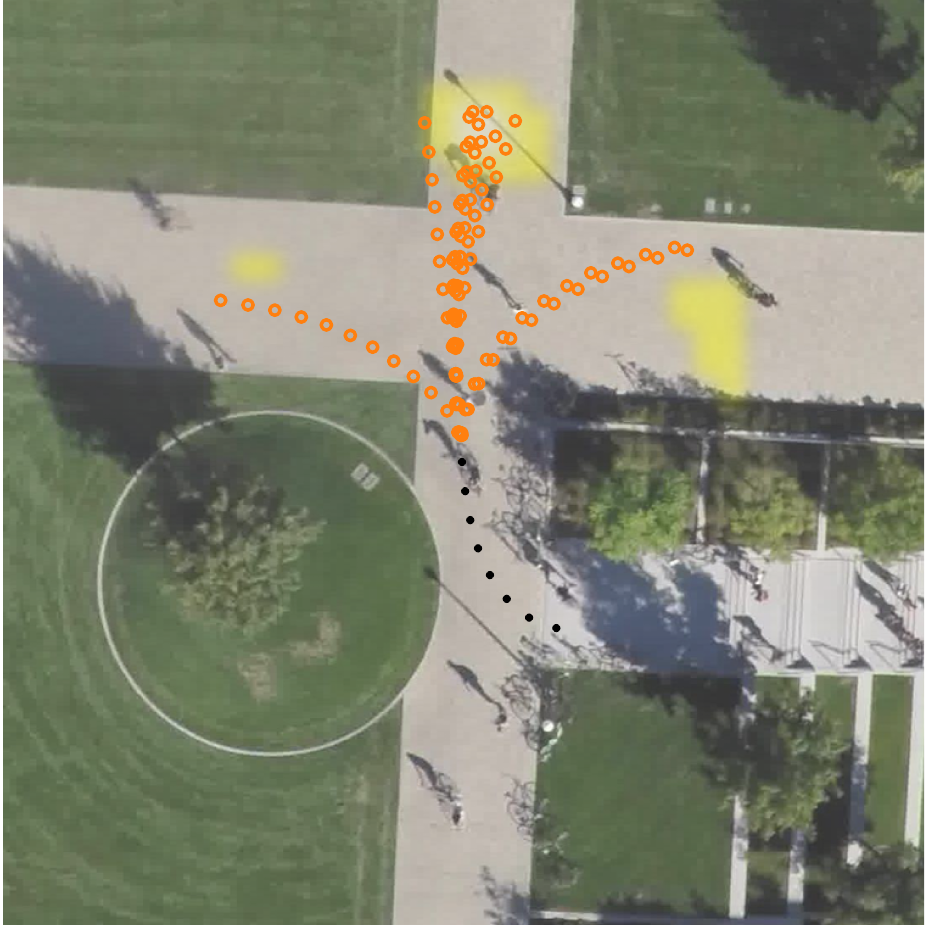}
\caption{Goal-GAN}
\vspace{2pt}
\end{subfigure}
\vspace{+0.2cm}
\begin{subfigure}{1\textwidth}
\includegraphics[width=0.23\textwidth]{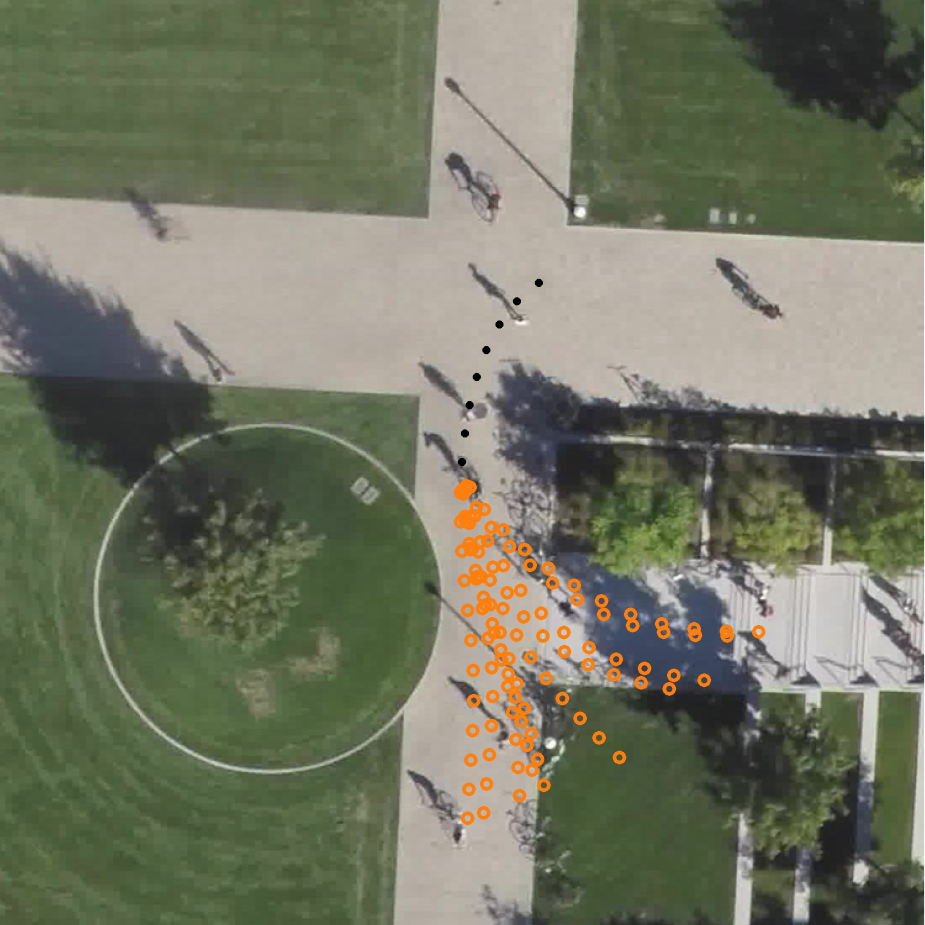}
\hfill
\includegraphics[width=0.23\textwidth]{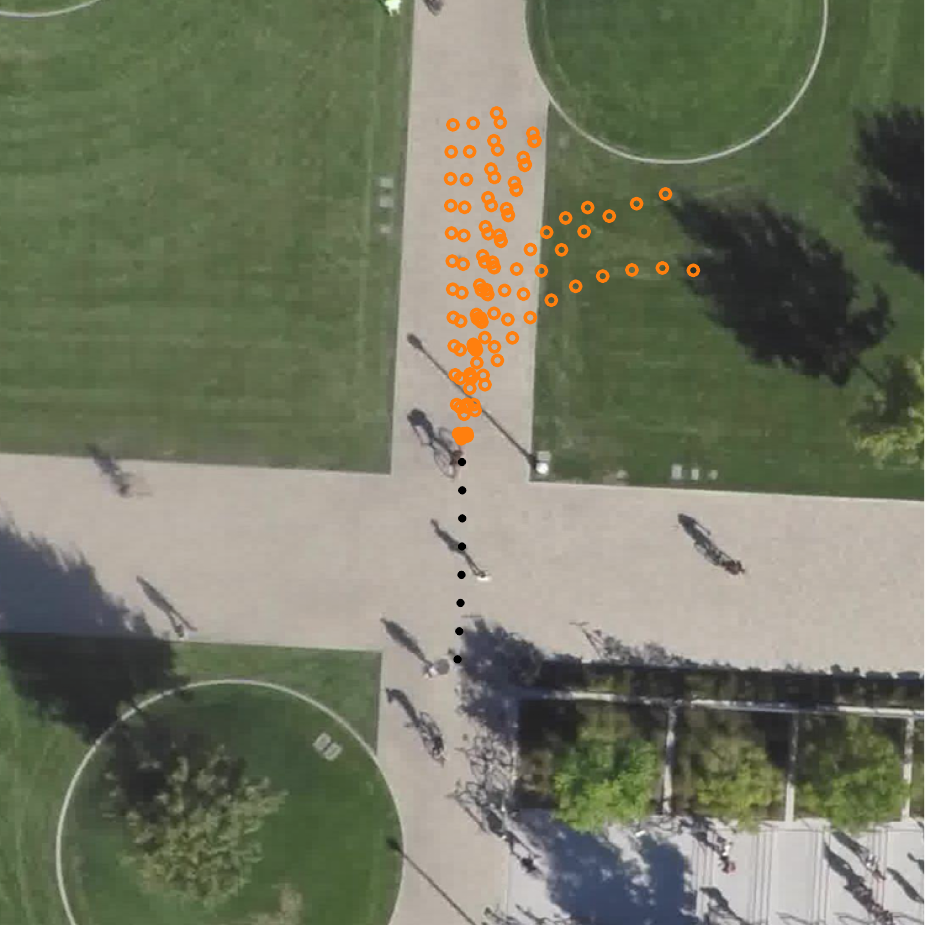}
\hfill
\includegraphics[width=0.23\textwidth]{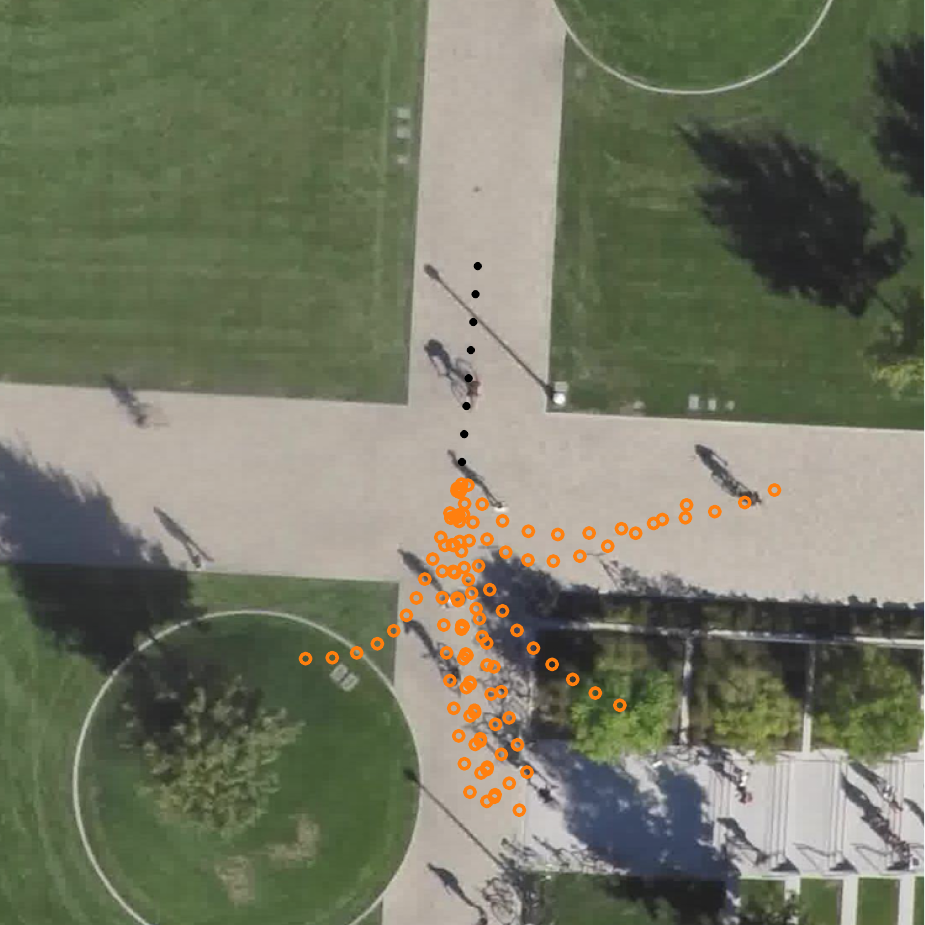}
\hfill
\includegraphics[width=0.23\textwidth]{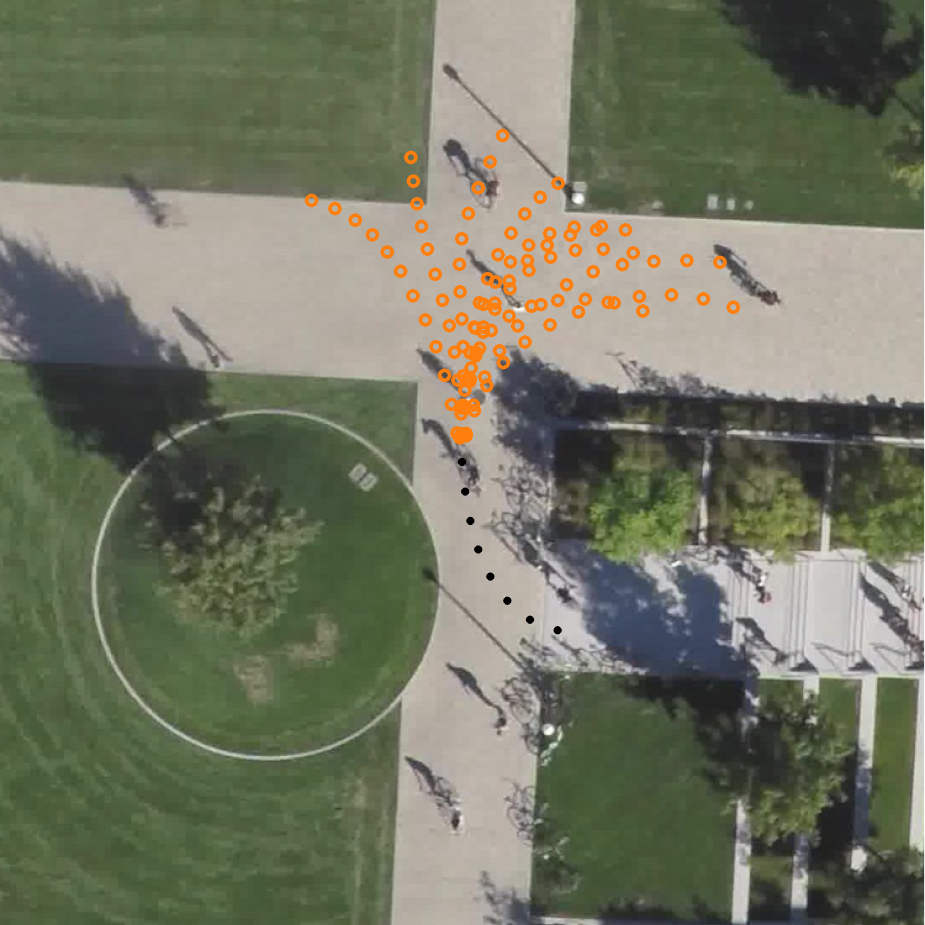}
\caption{Vanilla GAN}
\end{subfigure}

\caption{Visualisation of multiple generated trajectories (orange) for past trajectory (black) on the synthetic dataset. We compare the output of our Goal-GAN against the performance of the vanilla GAN using visual attention for $t_{pred}= 12$. For Goal-GAN, the yellow heatmap corresponds to the goal probability map.}\label{Fig:syntheticdata}
\end{figure}
%
\subsection{Qualitative Evaluation}
In this section, we visually inspect trajectories, generated by our model, and assess the quality of the predictions. 

\noindent {\bf Synthetic Dataset:} In \Cref{Fig:syntheticdata} we visualize trajectories of the synthetic dataset for our proposed Goal-GAN (top) and the vanilla GAN baseline~\cite{SGAN} (bottom).
Next to the predicted trajectories (orange circles), we display the probability distribution (yellow heatmap) of goal positions, estimated by the Goal Module. 
As shown in \Cref{Fig:syntheticdata}, Goal-GAN predicts a diverse set of trajectories routing to specific estimated modes. Here, we observe that Goal-GAN outputs an interpretable probability distribution that allows us to understand where the model ``sees`` the dominant modes in the scene. 
Comparing the quality of the predictions, we can demonstrate that Goal-GAN produces distinct modes while the GAN baseline tends to instead span its trajectory over a wider range leading to unfeasible paths. 

\begin{figure}[ht]
\begin{center}
\begin{subfigure}{0.25\textwidth}
\centering
\includegraphics[width=0.95\textwidth , height=0.9\textwidth]{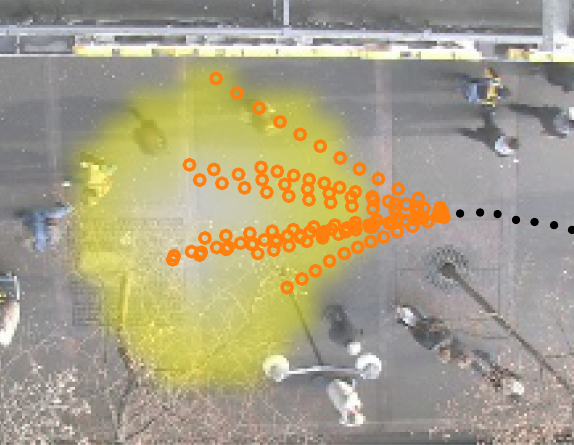}
\caption{ETH}\label{eth}
\end{subfigure}%
\begin{subfigure}{0.25\linewidth}
\centering
\includegraphics[width=0.95\textwidth, height=0.9\textwidth]{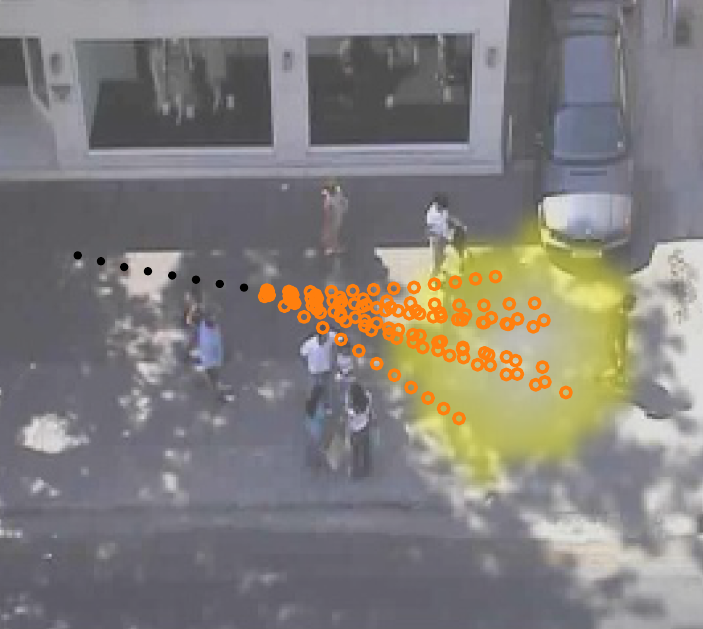}
\caption{Zara 2}\label{zara}
\end{subfigure}%
\begin{subfigure}{0.25\textwidth}
\centering
\includegraphics[width=0.95\textwidth , height=0.9\textwidth]{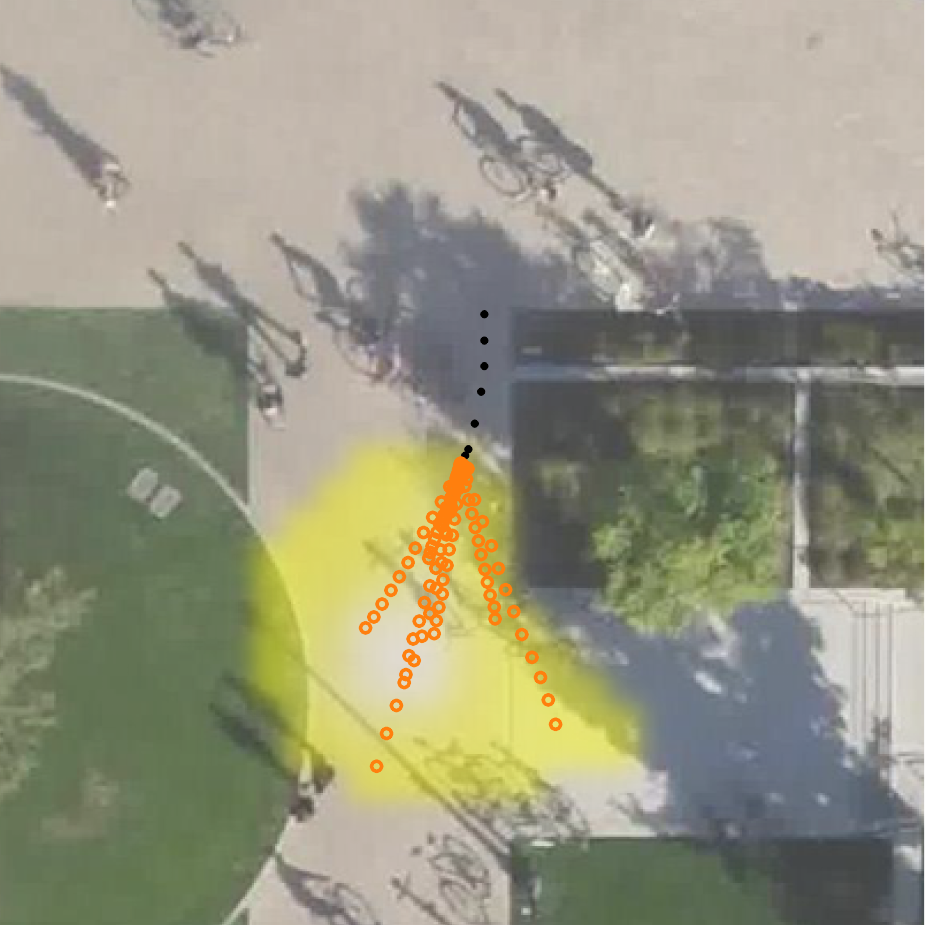}
\caption{Hyang 4 0}\label{little}
\end{subfigure}%
\begin{subfigure}{0.25\linewidth}
\centering
\includegraphics[width=0.95\textwidth ,
height=0.9\textwidth]{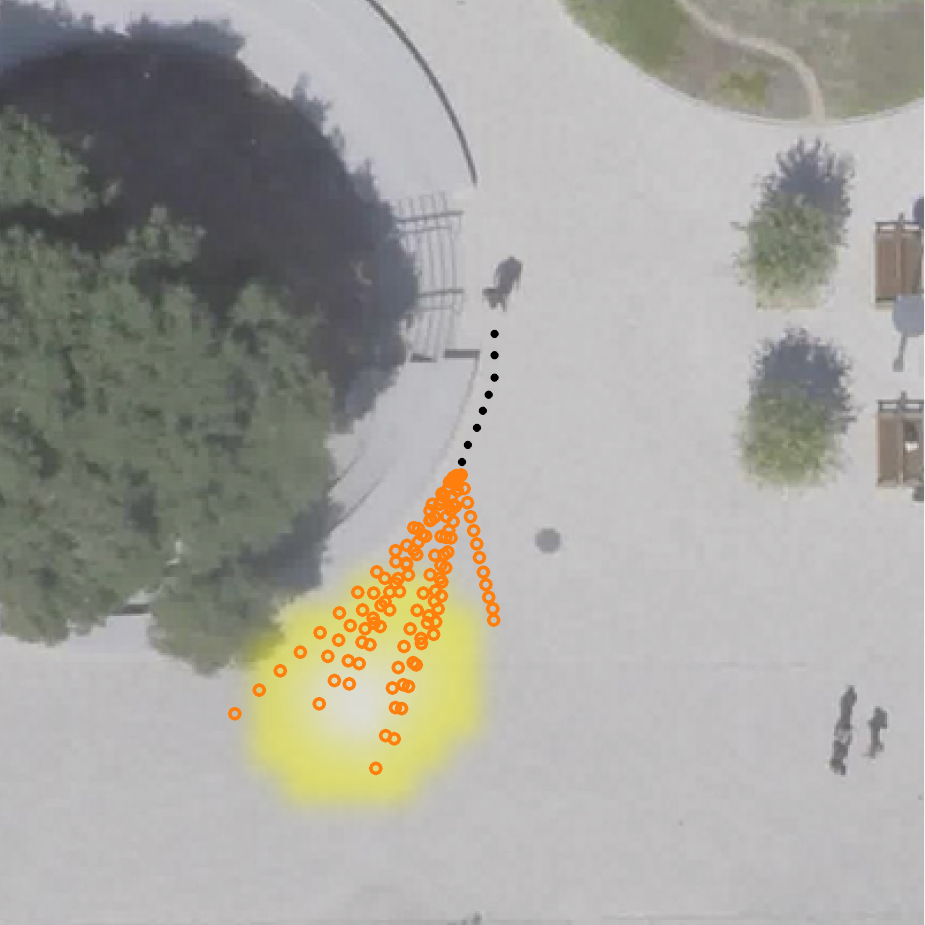}
\caption{Coupa 1}\label{coupa}
\end{subfigure}%
\end{center}
\caption{Visualisation of generated trajectories (orange circles) and estimated global goal probabilities (yellow heatmap).  The figures show that the model interacts with the visual context of the scene and ensures feasibility predictions.}\label{Fig:RealData}
\end{figure}
\noindent {\bf Real Data:}
Furthermore, we present qualitative results of the datasets ETH/UCY and SDD in \Cref{Fig:RealData}. 
The two figures show predictions on the \textit{Hotel} (\Cref{eth}) and \textit{Zara 2} (\Cref{zara}) sequences. Our model assigns high probability to a large area in the scene as in \textit{Hotel} sequence, as several positions could be plausible goals.  The broad distribution ensures that we generate diverse trajectories when there are no physical obstacles. Note that the generated trajectories do not only vary in direction but also in terms of speed.
In \textit{Zara 2}, the model recognizes the feasible area on the sidewalk and predicts no probability mass on the street or in the areas covered by the parked cars. 
In the scene \textit{Hyang 4} SDD dataset, we observe that the model successfully identifies that the pedestrian is walking on the path, assigning a very low goal probability to the areas, overgrown by the tree. 
This scenario is also presented successfully with synthetic data which shows that we can compare the results of the synthetic dataset to the behavior of real data. The trajectories shown for \textit{Coupa 1} demonstrate that the model generates solely paths onto concrete but avoids predictions leading towards the area of the tree.
\section{Conclusion}
In this work, we present Goal-GAN, a novel two-stage network for the task of pedestrian trajectory prediction. 
With the increasing interest in the interpretability of data-driven models, Goal-GAN allows us to comprehend the different stages iduring the prediction process. This is an alternative to the current generative models, which use a latent noise vector to encourage multimodality and diversity of the trajectory predictions. 
Our model achieves state-of-the-art results on the ETH, UCY, and SDD datasets while being able to generate multimodal, diverse, and feasible trajectories, as we experimentally demonstrate. 

\bibliographystyle{splncs}
\bibliography{egbib}

\end{document}